\documentclass[11pt]{article}

\usepackage[final]{acl}
\usepackage{times}
\usepackage{latexsym}
\usepackage[T1]{fontenc}
\usepackage[utf8]{inputenc}
\usepackage{microtype}
\usepackage{inconsolata}
\usepackage{graphicx}
\usepackage{subcaption}
\usepackage{ulem}
\usepackage{langsci-gb4e}
\usepackage{nnext}
\usepackage{leipzig}

\makeatletter
\ifacl@finalcopy
  \newcommand{\utku}[1]{\textcolor{black}{#1}}
  
\else
  \newcommand{\utku}[1]{\textcolor{red}{#1}}
  
\fi
\makeatother

\title{Quantifying the cross-linguistic effects of syncretism on agreement attraction}

\author{
  \textbf{Utku Turk} \\
  University of Maryland, College Park \\
  \texttt{utkuturk@umd.edu} \\
  \And
  \textbf{Eva Neu} \\
  University of Massachusetts, Amherst \\
  \texttt{eneu@umass.edu} \\
}

\begin{document}
\maketitle
\begin{abstract}

Agreement attraction errors, in which a verb erroneously agrees with an intervening noun rather than its grammatical head, are amplified by morphological syncretism in some languages (English, German, Russian) but not others (Turkish, Armenian), a cross-linguistic pattern without a principled account. We use surprisal and attention entropy from large language models as processing proxies to investigate this variation across four languages. LLM-derived measures replicate behavioral findings in English and German (syncretism modulates attraction), align with Turkish null results (no modulation), and partially capture Russian patterns. 
We discuss further directions for better understanding why syncretism affects agreement attraction differently across languages.

\end{abstract}

\section{Introduction}

\subsection{Agreement attraction and cue-based retrieval}

Consider the following sentences in \Next{}:\vspace{-0.5em}

\ea
    \ea[*]{\label{a}The key to the cabinet are rusty.}\vspace{-0.2em}
    \ex[*]{\label{b}The key to the cabinets are rusty.}\vspace{-0.5em}
    \z
\z

\noindent Both sentences are ungrammatical because the verb incorrectly bears plural rather than singular marking. Nonetheless, psycholinguistic research has shown that speakers are more likely to produce, and consider grammatical, sentences like \xref{b},  where the initial preamble includes a plural noun \cite{wagersetal2009}. These systematic errors are known as agreement attraction: instead of agreeing with the head noun---here, \textit{key}---, the verb erroneously agrees with an intervening attractor, \textit{cabinets}.

One theory that explains agreement attraction effects is the cue-based retrieval account \cite{wagersetal2009, dillonetal2013}. Following the ACT-R model of sentence processing, this model assumes that words are stored in a content-addressable memory as a chunk that can be later retrieved using specific cues \cite{lewisvasishth2005}. For instance, in \xref{b}, the noun \textit{key} might be stored as the chunk [SG, NOM, TP], encoding number, overt case marking and the dominating syntactic position. Similarly, the word \textit{cabinets} might be stored as [PL, NOM, PP].

According to the account put forward by \citet{wagersetal2009}, readers form a prediction about the form in which the verb should surface based on the chunks they have processed first. If the noun that occupies the syntactic position of an agreement controller [TP] is singular, they predict that the verb surfaces with singular agreement. In cases where participants indeed see a verb with singular marking such as \textit{is}, no disruption in reading or judgment is seen.\footnote{\utku{Other models of agreement attraction exist and make different predictions on processing grammatical sentences, most notably Marking and Morphing \cite{eberhard2005making, hammerly2019grammaticality}. We do not focus on these here; see \citet{yadav2023number} for an overview and computational comparison. These models are better suited to account for semantic integration \cite{solomon2004semantic}, as well as collectivity and distributivity effects \cite{humphreys2005notional} that are also known to modulate both agreement and agreement attraction.}}

On the other hand, if participants encounter a plural verb such as \textit{are} that does not match with the predicted number, a repair process is initiated. In this process, participants use the relevant cues provided by the verb to search for an agreement controller. The verb \textit{are}, which must agree with a plural noun phrase in subject position bearing nominative case, would trigger the search for a [PL, NOM, TP] chunk. In \xref{b}, no noun phrase matches this chunk perfectly. Accordingly, most of the time participants do not find these sentences grammatical. However, occasionally but systematically, participants erroneously retrieve \textit{cabinets}, encoded as [PL, NOM, PP], as the agreement controller due to a partial match with the cues provided by \textit{are}. In contrast, participants are much less likely to erroneously retrieve the word \textit{cabinet} in \xref{a}, since the features shared between the cues of \textit{are} [PL, NOM, TP] and the chunk of \textit{cabinet} [SG, NOM, PP] are even fewer.

\subsection{The effect of syncretism}

Two grammatical forms are said to be syncretic if they are realized with the same overt morphology despite bearing different syntactic and semantic features. E.g., most noun phrases in English---such as \textit{the dog}---are syncretic between nominative and accusative case marking. An exception are some pronouns which differ in their nominative and accusative forms, as with the first person pronoun \textit{I} (nominative) vs. \textit{me} (accusative).

Under the cue-based retrieval model, it has been argued that in the case of syncretism, a noun phrase is encoded with all possible features that could be realized by its morphology. E.g., \textit{the dog} would be encoded as bearing both NOM and ACC features despite the fact that it only ever bears one of them in the context of a specific sentence. In contrast, \textit{I} would be encoded with only NOM features, and \textit{me} with only ACC features. Given the broader assumptions about cue-based retrieval outlined above, this predicts that an attractor is more likely to interfere with the processing of subject-verb agreement if it is syncretic with the target.

By way of example, such an effect of syncretism on agreement attraction rates can be observed in a study by \citet{hartsuikeretal2003} on German. The German plural determiner surfaces as \textit{die} in both the nominative and the accusative case, but as a distinct form, \textit{den}, in the dative. Using the paradigm in (\ref{ga}-\ref{gd}), Hartsuiker et al. showed that participants made more agreement errors while completing sentences with syncretic determiners (\textit{die}).

    \ea\label{ga}
    \small
    \gll Die Stellungnahme zu den\textsubscript{DAT} Demonstrationen\\
    The position on the demonstrations\\
    \ex\label{gd}
    \small
    \gll Die Stellungnahme gegen die\textsubscript{ACC} Demonstrationen\\
    The position against the demonstrations\\
    \z

\noindent Under the cue-based retrieval model, this is as expected: since the determiner \textit{die} in (\ref{gd}) is encoded as both ACC and NOM, \textit{die Demonstrationen} is more likely to be misretrieved than the attractor in (\ref{ga}), \textit{den Demonstrationen}, which is only encoded as DAT.

\subsection{Cross-linguistic differences}

While an effect of syncretism on agreement attraction can be accounted for from the perspective of cue-based retrieval, the cross-linguistic picture is more complex. Syncretism has been shown to increase agreement attraction in English \cite{nicoletal2016}, Slovak \cite{badeckerkuminiak2007} and Czech \cite{lacinaetal2025}, besides German as seen above. However, in Turkish \cite{turklogacev2024} and Armenian \cite{avetisyanetal2020}, syncretism does not appear to affect error rates \utku{or reading times, respectively}. Moreover, French shows a reverse pattern, with syncretic attractors producing \textit{fewer} errors \cite{francketal2006}. Lastly, in addition to normal syncretism effects like English, Russian also shows interference from pseudo-plural attractors, which are singular, but syncretic with a plural form \cite{slioussar2018}.

To the best of our knowledge, this cross-linguistic variation in the effect of syncretism has yet to receive a principled account. One tentative explanation, suggested by \citet{dillonkeshev2024}, is that languages might differ as to how useful morphological cues are for resolving agreement. If speakers rely heavily on morphological marking to track agreement dependencies, error rates will highly depend on whether or not target and attractor are syncretic. On the other hand, if speakers rather rely on other factors such as word order or semantics to establish a relation between the verb and its subject, syncretism should have little effect. While plausible, this explanation still lacks empirical confirmation.

Here, we use LLMs as a cross-linguistically uniform processing proxy to quantify syncretism effects on agreement attraction across four languages (English, German, Russian, Turkish), extracting surprisal at the verb and attention entropy (i.e., how diffusely the model attends to candidate nouns at the verb) over potential agreement controllers. We find that LLM-derived measures replicate behavioral data in English, German, and Turkish, and partially in Russian. 
The one case where our LLM results are clearly at odds with the behavioral measures---Russian pseudoplurals \citep{slioussar2018}---suggests that humans are more easily led astray by superficial morphological similarity, whereas the LLMs draw more heavily on syntactic information from the broader context of the sentence which clearly disambiguates the morphological marking \utku{\citep[see also][for similar partial results]{bazhukov2024models}}.

\section{Methods}

A core finding in computational psycholinguistics is that the difficulty of processing a word, as measured by reading times, is proportional to its surprisal---the negative log probability of that word given its context \cite{hale2001,levy2008}. This relationship has been robustly demonstrated across reading paradigms: surprisal estimates from language models correlate linearly with self-paced reading times and eye-tracking measures such as gaze duration and total reading time \cite{smithlevy2013,goodkindbicknell2018}. Critically, surprisal from neural language models like GPT-2 has been shown to predict human reading behavior as well as or better than traditional n-gram models \cite{wilcoxetal2020,merkxfrank2021}.

This general correlation between reading times and surprisal values also applies to agreement attraction data. \citet{ryulewis2021} have argued that LLMs provide a correlate for the dynamics of memory retrieval according to the cue-based retrieval account, previously discussed in Section 1.1 (but  see also \citealp{arehallilinzen2020}). They showed that agreement attraction effects, as measured experimentally in reading times, can be replicated in LLMs using surprisal values, which quantify how likely a word is to appear in a context \cite{levy2008}. Concretely, in an ungrammatical sentence like \xref{b}, *\textit{The key to the cabinets are rusty}, lower surprisal at the verb corresponds to a stronger effect of agreement attraction since it indicates the possibility that the model is, like humans, led astray by the plural attractor and \textit{illusioned} into deeming the sentence grammatical. In terms of cue-based retrieval, decreased surprisal corresponds to the facilitation due to a partial match between the agreement bearer \textit{are} and the attractor \textit{cabinets}. This partial match would also predict an increased retrieval of the \textit{cabinets} as the agreement controller compared to the cases with singular attractor \xref{a}.

In addition to surprisal, Ryu and Lewis argue that behavioral agreement attraction data also correlate with attention values in an LLM which quantify the importance of a specific word in the context \cite{clarketal2019}. How much attention the model attributes, at the point of the verb, to a given noun indicates how likely it considers this noun to be the subject that is related to the verb by an agreement dependency. Accordingly, the more attention the model attributes  to the intervening attractor relative to the true target of agreement, the stronger the agreement attraction effect.

Ryu and Lewis's work has focused on linking attention and surprisal in LLMs specifically to reading time data on agreement attraction. However, in experimental studies, including some that we draw on in the present work, agreement attraction is often quantified using offline behavioral measures like acceptability judgments rather than reading times. Crucially, the memory-retrieval assumptions underlying Ryu and Lewis's framework are not paradigm-specific: cue-based retrieval makes predictions about processing difficulty in general, and retrieval interference effects are routinely observed across both online and offline measures in psycholinguistics \cite{wagersetal2009, dillonetal2013}. Some recent work further supports the offline--LLM link directly \cite{leeetal2024, timkeylinzen2023}, and we treat our measures as general-purpose proxies for retrieval competition rather than as direct models of any specific experimental paradigm. We note that some of the behavioral datasets we model come from production paradigms, while LLM-based measures are mainly used on the comprehension side \citep[cf.][]{harmonkapatsinski2021}. We hope to complement the current findings with self-paced reading and acceptability judgment studies to complete the cross-paradigm picture.

In the following, we describe our methodology for extracting surprisal and attention values more in detail.

\paragraph{Surprisal} We estimated surprisal for the agreement-bearing words (auxiliary or main verbs) using GPT-2-style autoregressive language models. Because surprisal is defined as the negative log probability of a word given its preceding context, autoregressive models that generate text left-to-right provide the most theoretically appropriate estimate.

\paragraph{Attention} We extracted attention weights from BERT-style bidirectional models, measuring the soft attention from the verb position to each noun that could either correctly (grammatical head) or erroneously (attractor) control agreement. Following methodological insights from probing studies \cite{voitaetal2019, clarketal2019}, we identified syntactically relevant attention heads rather than aggregating across all heads and layers.

Our approach adapts the entropy metric from \citet{ryulewis2021}, who aggregate attention entropy across syntactically selected heads spanning all layers. \citet{ohschuler2022} instead restrict computation to the final layer, but \citeauthor{ryulewis2021} note that none of \citet{ohschuler2022} selected syntactic heads fall in the final layer, suggesting that intermediate layers are more informative for grammatical dependencies. We follow this insight by using probing to select the most informative layer rather than defaulting to the final one, while keeping the layer-level aggregation of \citet{ohschuler2022}. Concretely, to identify which layer best tracks subject-verb dependencies, we parsed 1M-sentence corpora from the Leipzig Corpora Collection for each language using Universal Dependencies \cite{demarneffeetal2021} and selected the layer where the subject most reliably appeared among the verb's five most-attended tokens: layer 6 for English (63.1\% accuracy), layer 9 for German (48.2\%), layer 8 for Russian (69.3\%), and layer 8 for Turkish (53.0\%). Attention entropy was then computed over the mean attention distribution across all heads in that layer on the experimental stimuli.

In what follows, we focus on ungrammatical sentences and ask whether adding a plural attractor changes the measure relative to a singular attractor. \utku{We compare sentence pairs that only differ in the number marking of the attractor. Our measure of interest is} $\Delta = \mu_{\mathrm{Plural}} - \mu_{\mathrm{Singular}}$, computed separately for syncretic and non-syncretic conditions. A more negative $\Delta$ for surprisal means the plural attractor made the ungrammatical verb less surprising — i.e., stronger agreement attraction. A more positive $\Delta$ for attention entropy means attention was more dispersed, reflecting greater retrieval competition.

\section{Materials and results}

For inference, we fit per-language Bayesian mixed-effects models separately for surprisal and attention entropy. Each model includes all main effects and interactions among Syncretism, Grammaticality, and Attractor Number as fixed effects. The random-effects structure accounts for by-item variability. Items were allowed to vary in their intercepts as well as their slopes for Grammaticality, Attractor Number, and their interaction, with these random effects treated as uncorrelated. Fixed effects use treatment coding with reference levels Syncretic (Syncretism), Grammatical (Grammaticality), and Singular (Attractor). Under this coding, the two-way term $\beta_{G\times A}$ indexes the Grammaticality$\times$Attractor interaction in the Syncretic baseline, and the three-way term $\beta_{S\times G\times A}$ indexes how that interaction changes in the Non-syncretic condition (thus, the Non-syncretic interaction is $\beta_{G\times A}+\beta_{S\times G\times A}$). In the main text, we report posterior probabilities $P(\mathrm{effect}{>}0)$ for (i) the two-way Grammaticality$\times$Attractor interaction (attraction effect) and (ii) the three-way Syncretism$\times$Grammaticality$\times$Attractor interaction (syncretism modulation of attraction). Full model details and priors are reported in the Appendix.

\subsection{Experiment 1: German}

\begin{figure*}[t]
\centering
\begin{subfigure}[t]{0.44\linewidth}
    \centering
    \includegraphics[width=\linewidth]{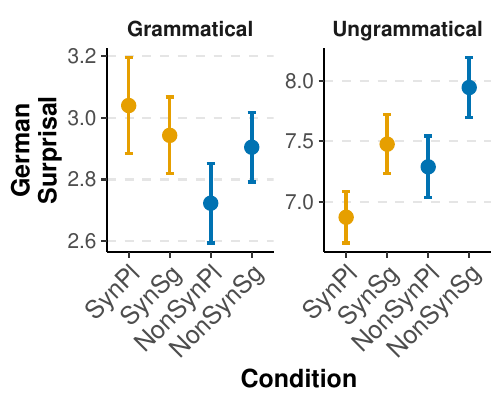}
    \label{fig:german-surprisal}
\end{subfigure}
\hfill
\begin{subfigure}[t]{0.44\linewidth}
    \centering
    \includegraphics[width=\linewidth]{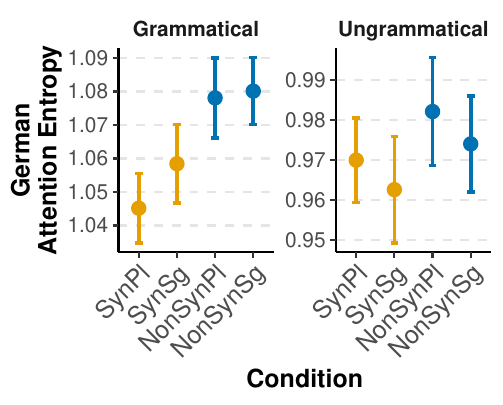}
    \label{fig:german-entropy}
\end{subfigure}
\vspace{-1.5em}
\caption{German model-derived surprisal and attention entropy measures (means with SE bars).}
\label{fig:german}
\end{figure*}

\paragraph{Materials} For German, we modeled the data by \citet{hartsuikeretal2003} summarized in Section 1.2. Recall that this  experiment found that participants made more agreement errors in production when completing sentences with accusative attractors that are syncretic with the nominative (marked with the determiner \textit{die}) compared to dative, non-syncretic attractors (marked with \textit{den}). We created grammatical (singular) and erroneous (plural) continuations for each experimental condition and modeled these data using \texttt{bert-base-german-cased} and \texttt{dbmdz/german-gpt2}. We expected to find overall lower surprisal and higher attention to the plural attractor in the ungrammatical completions, compared to singular attractor conditions. More importantly, we further predicted that this difference would be amplified in the syncretic condition \xref{ga} compared to the non-syncretic condition \xref{gd}.

\paragraph{Surprisal (Fig.\ref{fig:german})} Ungrammatical sentences showed higher surprisal compared to grammatical sentences as expected, and singular attractor conditions in ungrammatical sentences showed higher surprisal compared to plural attractors. The important comparison is whether the difference between plural and singular attractors is larger in syncretic than in non-syncretic items. The differences are very similar: non-syncretic $\Delta{=}-0.66$ (SE=0.35) and syncretic $\Delta{=}-0.60$ (SE=0.32), so the syncretism modulation in descriptive terms is small ($\Delta_{\mathrm{syn}}{-}\Delta_{\mathrm{non}}{=}0.06$). However, there seems to be an overall syncretism effects, such that non-syncretic nouns induced more surprisal overall. Posterior probabilities still show a clear attraction interaction in the syncretic baseline ($P(\beta_{G\times A}{>}0){<}0.01$), while the three-way term is positive but uncertain ($P(\beta_{S\times G\times A}{>}0){=}0.822$).

\paragraph{Attention Entropy (Fig.\ref{fig:german})} Entropy shows the same pattern but much weaker: non-syncretic $\Delta{=}0.008$ (SE=0.018) and syncretic $\Delta{=}0.007$ (SE=0.017), with near-zero descriptive modulation ($-0.001$). Posterior probabilities suggest a positive attraction interaction in the syncretic baseline ($P(\beta_{G\times A}{>}0){=}0.969$), but no reliable three-way interaction ($P(\beta_{S\times G\times A}{>}0){=}0.250$).

\paragraph{Discussion} Behaviorally, German is expected to show stronger attraction in syncretic (\textit{die}) than non-syncretic (\textit{den}) conditions. In our current model outputs, both attention and surprisal effects align with attraction predictions, but these values underpredict the syncretic-vs.-non-syncretic contrast.


\subsection{Experiment 2: English}

\begin{figure*}[t]
\centering
\begin{subfigure}[t]{0.44\linewidth}
    \centering
    \includegraphics[width=\linewidth]{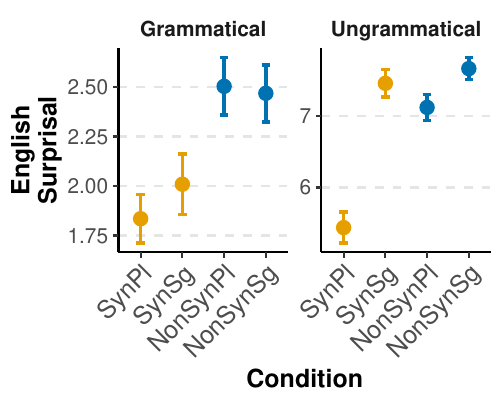}
    \label{fig:english-surprisal}
\end{subfigure}
\hfill
\begin{subfigure}[t]{0.44\linewidth}
    \centering
    \includegraphics[width=\linewidth]{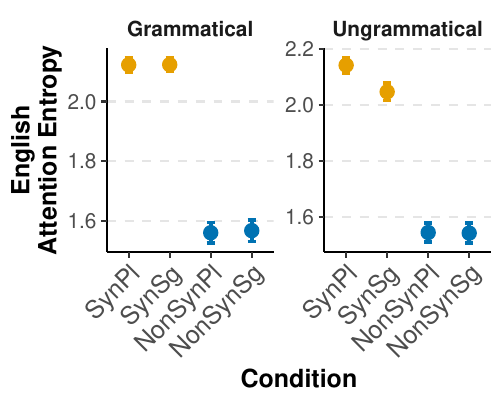}
    \label{fig:english-entropy}
\end{subfigure}
\vspace{-1.5em}
\caption{English model-derived surprisal and attention entropy measures (means with SE bars).}
\label{fig:english}
\end{figure*}

\paragraph{Materials} For English, we modeled data from \citet{nicoletal2016} and \citet{wagersetal2009} \utku{using pretrained GPT2-small \cite{radfordetal2019} and BERT  \cite{devlinetal2019a}}. \citet{wagersetal2009} found that participants were more likely to produce agreement attraction errors with plural attractors (e.g., \textit{cabinets}) compared to singular attractors (e.g., \textit{cabinet}). However, \utku{as noted above, nearly all English nouns are syncretic between nominative and accusative; thus,} Wagers et al.'s experiment was limited to attractors that were syncretic between nominative and accusative case marking. \citet{nicoletal2016} further investigated whether the syncretism of the attractor is crucial for agreement attraction effects. They compared accusative plural attractors whose case is syncretic with the nominative (e.g., \textit{gardens}) to genitive plural attractors which do not show this syncretism (e.g., \textit{elves'}) and found significantly higher interference rates with syncretic attractors. In their experiment, participants produced fewer agreement attraction errors with preambles like \xref{elfc}, which contain only a non-syncretic plural attractor (\textit{elves'}), than with \xref{elfb} or \xref{elfd}, which contain a syncretic plural attractor (\textit{gardens}). Moreover, there was no significant difference between \xref{elfb} and \xref{elfd}, indicating that the non-syncretic attractor did not increase interference rates when a syncretic attractor was present.

\ea
    \ea\label{elfa} The statue in the elf's garden \ldots{}\vspace{-0.2em}
    \ex\label{elfb} The statue in the elf's gardens \ldots{}\vspace{-0.2em}
    \ex\label{elfc} The statue in the elves' garden \ldots{}\vspace{-0.2em}
    \ex\label{elfd} The statue in the elves' gardens \ldots{}\vspace{-0.5em}
    \z
\z

\paragraph{Surprisal (Fig.\ref{fig:english})} For English, we focus on how much adding plural marking on a possible attractor (either \textit{elf} or \textit{garden}) changes processing in each morphology type: syncretic (\textit{elves' \underline{garden}} $\rightarrow$ \textit{elves' \underline{gardens}}) vs.\ non-syncretic (\textit{\underline{elf}'s gardens} $\rightarrow$ \textit{\underline{elves}' gardens}). Surprisal was sensitive to grammaticality as expected. Moreover, \utku{in ungrammatical items,} adding plural marking lowers surprisal much more in syncretic items ($\Delta{=}-2.009$, SE=0.286) than in non-syncretic items ($\Delta{=}-0.541$, SE=0.235). This appears as a strongly negative two-way interaction ($P(\beta_{G\times A}{>}0){<}0.01$) plus a strongly positive three-way interaction ($P(\beta_{S\times G\times A}{>}0){>}0.99$); i.e., the attraction pattern is clearly attenuated in non-syncretic items. 

\paragraph{Attention Entropy (Fig.\ref{fig:english})} Entropy shows the same descriptive direction: in ungrammatical items, plural increases entropy in syncretic items ($\Delta{=}0.095$, SE=0.042), but is near-zero in non-syncretic items ($\Delta{=}0.002$, SE=0.049). At model level, the syncretic-baseline two-way term is strongly positive ($P(\beta_{G\times A}{>}0){=}0.978$), while the three-way term is mostly negative ($P(\beta_{S\times G\times A}{>}0){=}0.090$), again consistent with attenuation in non-syncretic items.

\paragraph{Discussion} Research on English  has found higher agreement attraction rates with syncretic (accusative plural) than with non-syncretic attractors (genitive plural).  Our results from large language models correlate with these findings. Both surprisal and attention-related values showed a general effect of attraction results, but this effect was also modulated by the presence of syncretic nouns. The important difference in surprisal values increased with non-syncretic nouns, rendering non-syncretic plural meaningless in terms of facilitation effects. Similarly, the non-syncretic nouns also decreased attention entropy, meaning that the directed attention to the head was more concentrated, aligning with increased accuracy.


\subsection{Experiment 3: Russian}

\begin{figure*}[t]
\centering
\begin{subfigure}[t]{0.44\linewidth}
    \centering
    \includegraphics[width=\linewidth]{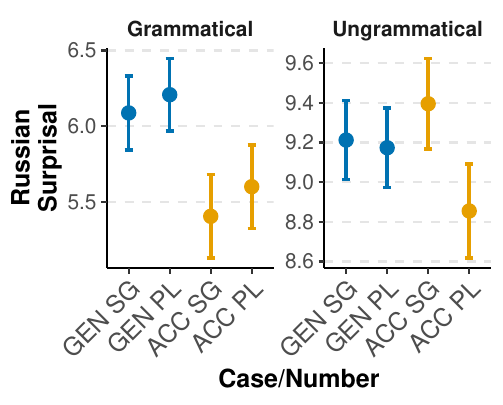}
    \label{fig:russian-surprisal}
\end{subfigure}
\hfill
\begin{subfigure}[t]{0.44\linewidth}
    \centering
    \includegraphics[width=\linewidth]{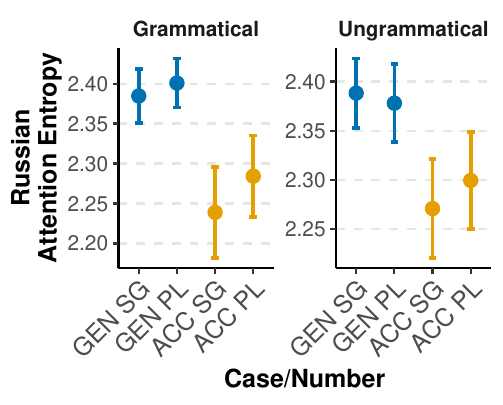}
    \label{fig:russian-entropy}
\end{subfigure}
\vspace{-1.5em}
\caption{Russian model-derived surprisal and attention entropy measures (means with SE bars).}
\label{fig:russian}
\end{figure*}

\paragraph{Materials} For Russian, we modeled data from \citet{slioussar2018} \utku{using \texttt{deepvk/bert-base-uncased} and \texttt{ai-forever/ rugpt3small\_based\_on\_gpt2}}. In some respect, Russian aligns with German and English in that speakers are more likely to wrongly consider a sentence with a singular head, a plural attractor and a plural verb grammatical if the case marking on the attractor is ambiguous. \citet{slioussar2018} showed this by comparing accusative attractors such as \textit{polja} (`field.\textsc{acc}.\textsc{pl}'), which are syncretic with their nominative form (\ref{russian-acc}), to genitive attractors such as \textit{ve\v{c}erinok} (`party.\textsc{gen}.\textsc{pl}'), which are not ambiguous with nominative marking (\ref{russian-gen}). They found that the syncretic attractors resulted in higher error rates.

\ea \label{russian-acc}\small
    \gll Trassa \v{c}erez polja/pole byli ... \\
    highway.\textsc{nom}.\textsc{sg} across field.\textsc{acc}.\textsc{pl}/\textsc{sg} were ...\\
    \glt \small `The highway across the fields/field were ...'
\ex \label{russian-gen} \small
    \gll Komnata dlja ve\v{c}erinok/ve\v{c}erinki byli ...\\
    room.\textsc{nom}.\textsc{sg} for party.\textsc{gen}.\textsc{pl}/\textsc{sg} were\\
    \glt \small `The room for parties/party were ...'
\z

However, Russian also exhibits another syncretism that affects agreement attraction rates, which differs from the patterns described so far. In (\ref{russian-gen}), the genitive singular form of `party', \textit{ve\v{c}erinki}, is syncretic with the nominative plural form, whereas the genitive plural form \textit{ve\v{c}erinok} does not exhibit a syncretism with the nominative. \utku{This pattern is widespread in Russian nominal paradigms, so any resulting effects cannot be attributed to rarity.} Slioussar found that with genitive attractors as in (\ref{russian-gen}), participants produced \textit{more} agreement attraction errors with singular than with plural attractors, contrary to the usual pattern. The effect was robust in both production and comprehension, and visible in reading times. What this suggests is that attractors can be misretrieved upon encountering a plural verb not only if they truly bear plural marking, but also if they are singular but syncretic with a plural form.

Following Slioussar's results, we expect to find a slightly different picture compared to English and German. We expect to see a two-way interaction that signals attraction effects in syncretic cases, and a three-way interaction showing a clear difference between syncretic and non-syncretic cases. However, we also expected to see a reversed attraction effect only in non-syncretic (genitive) cases, corresponding to higher interference rates from genitive singular than genitive plural attractors. To this end, we fitted an additional model only for the non-syncretic cases to see a non-interaction or reversed interaction.

\paragraph{Surprisal (Fig.\ref{fig:russian})} In Russian syncretic (ACC) ungrammatical items, adding a plural attractor lowers surprisal relative to singular ($\Delta_{\mathrm{Pl-Sg}}{=}-0.541$, SE=0.329). For this syncretic baseline case, the main model gives a strongly negative two-way interaction ($\beta_{G\times A}{=}-0.730$, $P(\beta_{G\times A}{>}0){<}0.01$). In non-syncretic (GEN) ungrammatical items, the descriptive contrast is much smaller and close to reversal: $\Delta_{\mathrm{Pl-Sg}}{=}-0.039$ (SE=0.282). The main-model three-way interaction is strongly positive ($\beta_{S\times G\times A}{=}0.568$, $P(\beta_{S\times G\times A}{>}0){>}0.99$), and the non-syncretic-only model gives a negative two-way interaction ($\beta_{G\times A}{=}-0.159$, $P(\beta_{G\times A}{>}0){<}0.01$), consistent with the reversal pattern, though the magnitude is very small compared to \citeauthor{slioussar2018}'s (\citeyear{slioussar2018}) results.

\paragraph{Attention entropy (Fig.\ref{fig:russian})} In Russian syncretic (ACC) ungrammatical items, adding a plural attractor slightly increases entropy relative to singular ($\Delta_{\mathrm{Pl-Sg}}{=}0.029$, SE=0.070). For this syncretic baseline case, the main-model two-way term is weak and not clearly directional ($\beta_{G\times A}{=}-0.016$, $P(\beta_{G\times A}{>}0){=}0.183$). In non-syncretic (GEN) ungrammatical items, the descriptive effect is reversed and near zero ($\Delta_{\mathrm{Pl-Sg}}{=}-0.010$, SE=0.053; equivalently $\Delta_{\mathrm{Sg-Pl}}{=}+0.010$). The main-model three-way interaction is also weak ($\beta_{S\times G\times A}{=}-0.010$, $P(\beta_{S\times G\times A}{>}0){=}0.352$), and the non-syncretic-only two-way term is similarly weak ($\beta_{G\times A}{=}-0.025$, $P(\beta_{G\times A}{>}0){=}0.138$).

\paragraph{Discussion} Our LLM based values show clear attraction effect within syncretic items. In non-syncretic cases, we find that it is heavily attenuated, similar to the English findings. While this overall effect is expected, Russian behavioral data also show a reversed effect within the GEN conditions in which the singular attractor leads to higher interference rates. However, this reversal is not fully supported, with much weaker non-syncretic (GEN) contrasts and weak non-syncretic-only reversed-attraction effects. Entropy remains less decisive overall.

\subsection{Experiment 4: Turkish}

\begin{figure*}[t]
\centering
\begin{subfigure}[t]{0.44\linewidth}
    \centering
    \includegraphics[width=\linewidth]{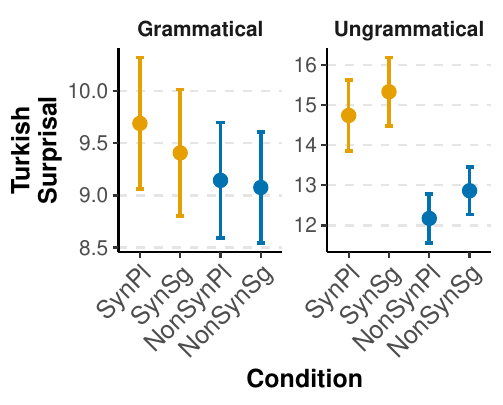}
    \label{fig:turkish-surprisal}
\end{subfigure}
\hfill
\begin{subfigure}[t]{0.44\linewidth}
    \centering
    \includegraphics[width=\linewidth]{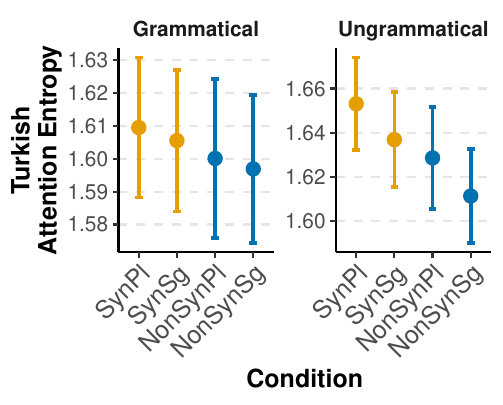}
    \label{fig:turkish-entropy}
\end{subfigure}
\vspace{-1.5em}
\caption{Turkish model-derived surprisal and attention entropy measures (means with SE bars).}
\label{fig:turkish}
\end{figure*}

\paragraph{Materials}

For Turkish, we modeled data from \citet{turklogacev2024}, who compared agreement attraction rates between sentences such as \xref{lago} and \xref{turk}, building on work by \citet{lagoetal2019}. \utku{We used \texttt{dbmdz/bert-base-turkish-128k-cased} and \texttt{redrussianarmy/gpt2-turkish-cased}}. The head noun (here, \textit{e\u{g}itmen-i} `instructor' or \textit{hoca-s{\i}} `teacher') bears a possessor morpheme and is preceded by a genitive-marked possessee (here, \textit{teknisyen-(ler)-in} `of the technician(s)').

\ea \label{lago} \small
\gll Teknisyen-(ler)-in e\u{g}itmen-i \ldots{} ko\c{s}tu-(lar).\\
tech-(\Pl{})-\Gen{} instructor-\Poss{} ~ ran-(\Pl{})\\
\glt \small`The instructor of the technician(s) \ldots{} ran$_{\Sg{}/\Pl{}}$.'
\ex \label{turk} \small
\gll Teknisyen-(ler)-in hoca-s{\i} \ldots{} ko\c{s}tu-(lar).\\
tech-(\Pl{})-\Gen{} teacher-\Poss{} ~ ran-(\Pl{})\\
\glt \small`The teacher of the technician(s) \ldots{} ran$_{\Sg{}/\Pl{}}$.'
\z

\noindent Since genitive case also surfaces on the subjects of embedded clauses in Turkish, speakers might misinterpret the possessee as an embedded subject and, if it is plural, misretrieve it when encountering a plural verb. Crucially, after a consonant as in \xref{lago}, the possessor morpheme on the head is realized as -\textit{I}, which is syncretic with accusative case. Vowel-final nouns as in (\ref{turk}), on the other hand, take the possessor morpheme -\textit{sI}, which is not syncretic but can only surface on nominative subjects.\footnote{Note that in the Turkish materials, the attractor precedes the head noun, whereas in the English, German, and Russian materials, the attractor intervenes between the head and the verb. At least for Russian and German, and possibly Turkish, alternative orders can be constructed (e.g., via topicalization or focus movement) without changing the structure drastically. \citet{sturt2024agreement} found that distractor position modulated the size of attraction effects in English; however, whether similar positional effects hold across languages and whether LLMs mirror human behavior under these rarer configurations remains an open question.} \citet{turklogacev2024} expected that the unambiguous nominative case marking on the head as in \xref{turk} (\textit{hoca-s{\i}}) would diminish interference rates from the genitive attractor \textit{teknisyen-ler-in}, but this prediction was not borne out. Unlike in the languages seen so far, syncretism does not appear to affect agreement attraction rates in Turkish.

Following their results, we expect that plural attractors overall decrease surprisal and increase attention entropy in ungrammatical sentences. However, this effect should not be modulated by syncretism.

\paragraph{Surprisal (Fig.\ref{fig:turkish})} For Turkish, the key test is whether plural-attractor effects differ by syncretism. They do not: adding a plural attractor lowers surprisal in both conditions, with the effects being of very similar size (non-syncretic $\Delta{=}-0.691$, SE=0.853; syncretic $\Delta{=}-0.590$, SE=1.230). Bayesian estimates show weak directional evidence for the syncretic-baseline two-way term ($P(\beta_{G\times A}{>}0){=}0.194$), and more importantly, no evidence for a three-way interaction ($P(\beta_{S\times G\times A}{>}0){=}0.332$).

\paragraph{Attention entropy (Fig.\ref{fig:turkish})} Entropy shows the same qualitative conclusion. The plural-attractor increase is almost identical in the two conditions (non-syncretic $\Delta{=}0.017$, SE=0.031; syncretic $\Delta{=}0.016$, SE=0.030). However, posterior probabilities show no directional evidence for both interaction terms ($P(\beta_{G\times A}{>}0){=}0.782$, $P(\beta_{S\times G\times A}{>}0){=}0.550$).

\paragraph{Discussion}

While our surprisal results align with the behavioral Turkish data---namely, less surprisal in agreement attraction cases---the attention values are inconclusive.  As for the syncretism, we show that it does not modulate attraction related facilitation within surprisal, and plural-attractor effects are comparable in syncretic and non-syncretic conditions. Given that there is a general sensitivity in LLM measures to attraction effects, the lack of syncretism modulation is consistent with the behavioral data.

\section{General discussion}

This study aimed to establish a baseline for future work on syncretism and morphological encoding in memory retrieval by identifying correlates between human behavioral data and LLM-based measures, following recent methodological advances \citep{wilcoxetal2020, ryulewis2021}. Across three of the four languages examined, model-derived measures matched the effects attested in the behavioral literature. Surprisal provided a clearer and more consistent signal, while attention entropy was less decisive: the expected patterns were present but weaker and less certain. This divergence is not unexpected, given that attention-based measures are more indirect and less established than surprisal as proxies for retrieval competition, but it points to room for improvement in how subject-tracking attention is extracted and interpreted.

The one clear exception is Russian. Surprisal successfully captured the standard attraction effect in the accusative (syncretic) condition, but failed to reproduce the reversed pattern in the genitive condition, where behavioral data show that a genitive singular attractor---syncretic with the nominative plural---generates more interference than a genuine genitive plural. One possible explanation is that LLMs are less sensitive than human comprehenders to spurious morphological overlaps between two forms that share no underlying feature ([GEN, SG] and [NOM, PL]) and/or rely more heavily on syntactic information to disambiguate these forms. Resolving this question requires targeted probing of how models represent syncretism with and without feature overlap, which we leave for future work.

More broadly, we hope that in the long term, this study contributes to understanding of \textit{why} syncretism modulates agreement attraction differently across languages. Our results suggest that LLMs are sensitive to the same syncretisms that drive attraction in human experiments, even if through different underlying mechanisms. A promising direction for future work is to ask whether languages differ systematically in how much weight they place on morphological cues when tracking dependencies. Syncretism effects may be strongest in languages where morphological distinctions are few but highly informative (English, German) or where case encodes a dense bundle of features (fusional languages like Russian and Czech). In agglutinative languages like Turkish and Armenian, where morphological information is more transparent and compositional, speakers and models alike may rely less on the presence or absence of a specific syncretic overlap, leading to the null effects observed here.

\appendix

\section{Model Specification and Full Results}

All Bayesian models were fit with the \texttt{brms} package in R (Gaussian family). Priors: $\mathrm{Normal}(0,2)$ on all fixed effects, $\mathrm{Student}\text{-}t(3,0,2.5)$ on the intercept, $\mathrm{Exponential}(1)$ on random-effect SDs and residual $\sigma$. Sampling: 4 chains $\times$ 12{,}000 iterations (2000 warmup), \texttt{adapt\_delta}${=}0.95$, \texttt{max\_treedepth}${=}12$, seed 1234. The main model formula was:
\begin{center}
\small\texttt{value $\sim$ Syn $*$ Gram $*$ Attr + (1 + Gram $*$ Attr\,||\,Item)}
\end{center}
Treatment coding: Syncretic, Grammatical, and Singular as reference levels, so the intercept represents the expected value for grammatical, syncretic, singular-attractor items. All Russian fits were restricted to singular-head items. Code and data for our paper can be found at \url{www.github.com/utkuturk/scil_TurkNeu2026}.

\begin{table*}[ht]
\centering
\footnotesize
\setlength{\tabcolsep}{3pt}
\caption{Full posterior summaries for the \textbf{Surprisal} models. Terms are shown as rows; languages as column groups. $\hat\beta$ is the posterior mean and $P$ is $P(\mathrm{effect}{>}0)$. $S$ = Syncretism (Non-syncretic vs.\ Syncretic), $G$ = Grammaticality (Ungrammatical vs.\ Grammatical), $A$ = Attractor (Plural vs.\ Singular). \textbf{Bold}: $P{>}0.89$ or $P{<}0.11$.}
\label{tab:brms-surprisal}
\begin{tabular}{l rr rr rr rr}
\hline
 & \multicolumn{2}{c}{English} & \multicolumn{2}{c}{German} & \multicolumn{2}{c}{Russian} & \multicolumn{2}{c}{Turkish} \\
\cline{2-3}\cline{4-5}\cline{6-7}\cline{8-9}
Term & $\hat\beta$ & $P$ & $\hat\beta$ & $P$ & $\hat\beta$ & $P$ & $\hat\beta$ & $P$ \\
\hline
Intercept         & $\phantom{-}2.091$ & $\mathbf{>0.999}$ & $\phantom{-}2.944$ & $\mathbf{>0.999}$ & $\phantom{-}5.411$ & $\mathbf{>0.999}$ & $\phantom{-}9.793$ & $\mathbf{>0.999}$ \\
$S$               & $\phantom{-}0.351$ & $\mathbf{0.972}$  & $-0.044$           & $0.339$           & $\phantom{-}0.611$ & $\mathbf{0.953}$  & $-0.795$           & $\mathbf{0.015}$  \\
$G$               & $\phantom{-}5.358$ & $\mathbf{>0.999}$ & $\phantom{-}4.460$ & $\mathbf{>0.999}$ & $\phantom{-}3.917$ & $\mathbf{>0.999}$ & $\phantom{-}5.481$ & $\mathbf{>0.999}$ \\
$A$               & $-0.220$           & $\mathbf{0.092}$  & $\phantom{-}0.085$ & $0.718$           & $\phantom{-}0.191$ & $\mathbf{>0.999}$ & $\phantom{-}0.099$ & $0.596$           \\
$S{\times}G$      & $-0.160$           & $0.244$           & $\phantom{-}0.512$ & $\mathbf{>0.999}$ & $-0.803$           & $\mathbf{0.009}$  & $-1.739$           & $\mathbf{<0.001}$ \\
$S{\times}A$      & $\phantom{-}0.270$ & $0.885$           & $-0.272$           & $\mathbf{0.053}$  & $-0.071$           & $0.207$           & $-0.022$           & $0.482$           \\
$G{\times}A$      & $-1.731$           & $\mathbf{<0.001}$ & $-0.690$           & $\mathbf{<0.001}$ & $-0.730$           & $\mathbf{<0.001}$ & $-0.488$           & $0.194$           \\
$S{\times}G{\times}A$ & $\phantom{-}1.131$ & $\mathbf{>0.999}$ & $\phantom{-}0.218$ & $0.822$       & $\phantom{-}0.568$ & $\mathbf{>0.999}$ & $-0.307$           & $0.332$           \\
\hline
\end{tabular}
\end{table*}

\begin{table*}[ht]
\centering
\footnotesize
\setlength{\tabcolsep}{3pt}
\caption{Full posterior summaries for the \textbf{Attention Entropy} models. Terms are shown as rows; languages as column groups. $\hat\beta$ is the posterior mean and $P$ is proportion of posterior samples that are bigger than 0 $P(\mathrm{effect}{>}0)$. $S$ = Syncretism (Non-syncretic vs.\ Syncretic), $G$ = Grammaticality (Ungrammatical vs.\ Grammatical), $A$ = Attractor (Plural vs.\ Singular). \textbf{Bold}: $P{>}0.89$ or $P{<}0.11$}
\label{tab:brms-entropy}
\begin{tabular}{l rr rr rr rr}
\hline
 & \multicolumn{2}{c}{English} & \multicolumn{2}{c}{German} & \multicolumn{2}{c}{Russian} & \multicolumn{2}{c}{Turkish} \\
\cline{2-3}\cline{4-5}\cline{6-7}\cline{8-9}
Term & $\hat\beta$ & $P$ & $\hat\beta$ & $P$ & $\hat\beta$ & $P$ & $\hat\beta$ & $P$ \\
\hline
Intercept         & $\phantom{-}2.106$ & $\mathbf{>0.999}$ & $\phantom{-}1.059$ & $\mathbf{>0.999}$ & $\phantom{-}2.237$ & $\mathbf{>0.999}$ & $\phantom{-}1.597$ & $\mathbf{>0.999}$ \\
$S$               & $-0.537$           & $\mathbf{<0.001}$ & $\phantom{-}0.022$ & $\mathbf{0.994}$  & $\phantom{-}0.147$ & $\mathbf{0.987}$  & $-0.006$           & $0.281$           \\
$G$               & $-0.076$           & $\mathbf{0.012}$  & $-0.096$           & $\mathbf{<0.001}$ & $\phantom{-}0.032$ & $\mathbf{0.956}$  & $\phantom{-}0.031$ & $\mathbf{>0.999}$ \\
$A$               & $\phantom{-}0.000$ & $0.513$           & $-0.013$           & $\mathbf{0.074}$  & $\phantom{-}0.045$ & $\mathbf{0.988}$  & $\phantom{-}0.004$ & $0.643$           \\
$S{\times}G$      & $\phantom{-}0.049$ & $0.867$           & $-0.011$           & $0.163$           & $-0.027$           & $0.166$           & $-0.018$           & $0.120$           \\
$S{\times}A$      & $-0.008$           & $0.411$           & $\phantom{-}0.011$ & $0.831$           & $-0.029$           & $0.150$           & $-0.001$           & $0.471$           \\
$G{\times}A$      & $\phantom{-}0.093$ & $\mathbf{0.978}$  & $\phantom{-}0.021$ & $\mathbf{0.969}$  & $-0.016$           & $0.183$           & $\phantom{-}0.012$ & $0.782$           \\
$S{\times}G{\times}A$ & $-0.083$       & $\mathbf{0.090}$  & $-0.010$           & $0.250$           & $-0.010$           & $0.352$           & $\phantom{-}0.002$ & $0.550$           \\
\hline
\end{tabular}
\end{table*}

\begin{table*}[ht]
\centering
\footnotesize
\setlength{\tabcolsep}{3pt}
\caption{Russian non-syncretic-only models (GEN condition only). No Syncretism factor; intercept is the mean for grammatical, singular-attractor items. $\hat\beta$ is the posterior mean and $P$ is proportion of posterior samples that are bigger than 0 $P(\mathrm{effect}{>}0)$. $G$ = Grammaticality (Ungrammatical vs.\ Grammatical), $A$ = Attractor (Plural vs.\ Singular). \textbf{Bold}: $P{>}0.89$ or $P{<}0.11$}
\label{tab:brms-russian-nonsyn}
\begin{tabular}{l rr rr}
\hline
 & \multicolumn{2}{c}{Surprisal} & \multicolumn{2}{c}{Entropy} \\
\cline{2-3}\cline{4-5}
Term & $\hat\beta$ & $P$ & $\hat\beta$ & $P$ \\
\hline
Intercept    & $\phantom{-}6.054$ & $\mathbf{>0.999}$ & $\phantom{-}2.380$ & $\mathbf{>0.999}$ \\
$G$          & $\phantom{-}3.119$ & $\mathbf{>0.999}$ & $\phantom{-}0.004$ & $0.613$           \\
$A$          & $\phantom{-}0.117$ & $\mathbf{0.976}$  & $\phantom{-}0.016$ & $0.821$           \\
$G{\times}A$ & $-0.159$           & $\mathbf{<0.001}$ & $-0.025$           & $0.138$           \\
\hline
\end{tabular}
\end{table*}

\end{document}